\documentclass[journal]{IEEEtran}
\usepackage{amsmath,amsfonts}
\usepackage{amssymb}
\usepackage{algorithmic}
\usepackage[ruled,linesnumbered]{algorithm2e}
\usepackage{color}
\usepackage{bm}
\usepackage{bbm}

\usepackage{array}
\usepackage[caption=false,font=normalsize,labelfont=sf,textfont=sf]{subfig}
\usepackage{textcomp}
\usepackage{stfloats}
\usepackage{url}
\usepackage{verbatim}
\usepackage{graphicx}
\usepackage{cite}
\usepackage{threeparttable}

\usepackage{color}
\usepackage{multirow}
\usepackage{multicol}
\usepackage{marvosym}

\usepackage{doi}
\begin{document}

\title{Progressive Transfer Learning for Dexterous In-Hand Manipulation with Multi-Fingered Anthropomorphic Hand}



\author{Yongkang Luo\textsuperscript{1},~\IEEEmembership{Member,~IEEE}, Wanyi Li\textsuperscript{1},~\IEEEmembership{Member,~IEEE}, Peng Wang\textsuperscript{1, 2, 3},~\IEEEmembership{Member,~IEEE}, Haonan Duan\textsuperscript{1,2}, Wei Wei\textsuperscript{1,2},~\IEEEmembership{Student Member,~IEEE}, and Jia Sun\textsuperscript{1}
\thanks{This work was supported in part by the National Natural Science Foundation of China under Grants (91748131, 62006229, and 61771471), in part by the InnoHK Project, and in part by the Strategic Priority Research Program of Chinese Academy of Science under Grant XDB32050100. \itshape{(Corresponding author: Peng Wang.)}}
\thanks{E-mail: \{yongkang.luo, wanyi.li, peng\_wang, duanhaonan2021, wei.wei2018, jia.sun\}@ia.ac.cn.}
\thanks{$^{1}$ State Key Laboratory of Multimodal Artificial Intelligence Systems, Institute of Automation, Chinese Academy of Sciences, Beijing, 100190, China.} 
\thanks{$^{2}$ School of Artificial Intelligence, University of Chinese Academy of Sciences, Beijing, 100190, China}
\thanks{$^{3}$ Centre for Artificial Intelligence and Robotics, Hong Kong Institute of Science and Innovation, Chinese Academy of Sciences, Hong Kong, 999077, China}
}

\markboth{IEEE Transactions on Neural Networks and Learning Systems, 2023}%
{Shell \MakeLowercase{\textit{et al.}}: A Sample Article Using IEEEtran.cls for IEEE Journals}


\maketitle

\begin{abstract}

Dexterous in-hand manipulation for a multi-fingered anthropomorphic hand is extremely difficult because of the high-dimensional state and action spaces, rich contact patterns between the fingers and objects. Even though deep reinforcement learning has made moderate progress and demonstrated its strong potential for manipulation, it is still faced with certain challenges, such as large-scale data collection and high sample complexity. Especially, for some slight change scenes, it always needs to re-collect vast amounts of data and carry out numerous iterations of fine-tuning. Remarkably, humans can quickly transfer learned manipulation skills to different scenarios with little supervision. Inspired by human flexible transfer learning capability, we propose a novel dexterous in-hand manipulation progressive transfer learning framework (PTL) based on efficiently utilizing the collected trajectories and the source-trained dynamics model. This framework adopts progressive neural networks for dynamics model transfer learning on samples selected by a new samples selection method based on dynamics properties, rewards and scores of the trajectories. Experimental results on contact-rich anthropomorphic hand manipulation tasks show that our method can efficiently and effectively learn in-hand manipulation skills with a few online attempts and adjustment learning under the new scene. Compared to learning from scratch, our method can reduce training time costs by 95\%.

\end{abstract}


\section{Introduction}
\IEEEPARstart{D}{exterous} manipulation of objects is a fundamental task in human daily life, it is still one type of the most complex and largely unsolved control problems in robotics. The goal of dexterous in-hand manipulation is to control object movement with precise control of forces and motions for a multi-fingered anthropomorphic hand that has high degrees of freedom \cite{okamura2000overview}. Dexterous in-hand manipulation not only needs to effectively coordinate the high degree-of-freedom fingers, but also stabilize objects through contacts \cite{kumar2014real}.

As dexterous manipulation has been an active area of research for decades, many approaches have been proposed \cite{bicchi2000hands, ma2011dexterity,huang2000mechanics, erdmann1998exploration,bai2014dexterous}. These methods use open-loop planning after computing a trajectory, which is based on exact models of the robot hand and objects. Since accurate models are not easy to obtain, it is arduous for such methods to realize satisfactory performance. Some methods adopt a closed-loop approach for dexterous manipulation with sensor feedback during execution \cite{tahara2010dynamic, li2014learning}. These methods can correct the failure cases at runtime, but they still need reasonable models about the  kinematics and dynamics of the robot, which are difficult for an under-actuated multi-fingered anthropomorphic hand with high degrees of freedom.

\begin{figure*}[htbp]
\centering
\includegraphics[width=6.0in]{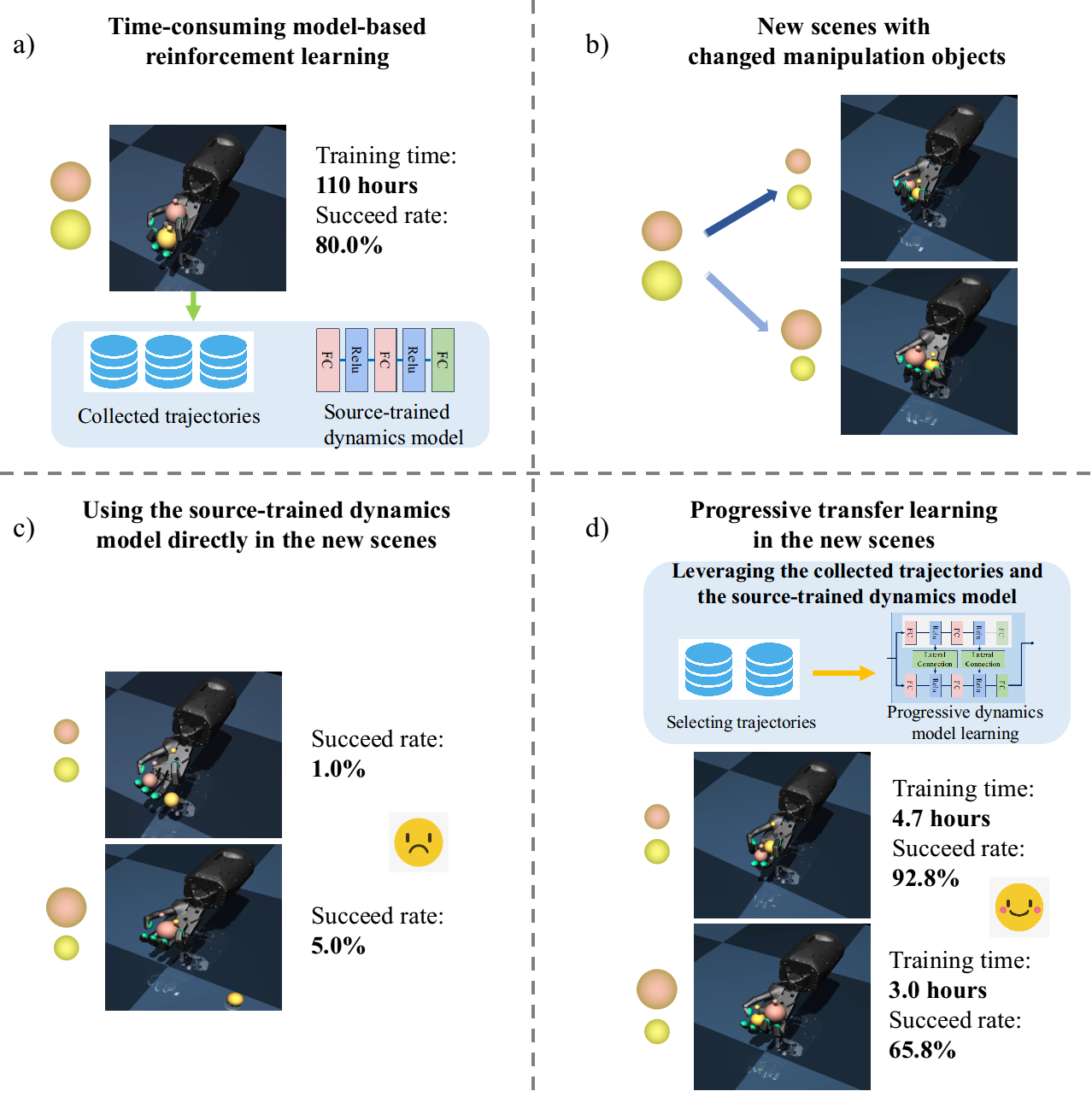}
\caption{Motivation of progressive transfer learning for in-hand manipulation with multi-fingered anthropomorphic hand. a) Manipulation learning based on model-based reinforcement learning is time-consuming. b) New scenes with changed manipulation objects. c) Using the source-trained model directly in the new scenes often fails. d) Progressive transfer learning in the new scenes through leveraging the collected trajectories and the source-trained dynamics model performs well.}
\label{motivation}
\end{figure*}

The development of deep reinforcement learning \cite{mnih2015human, silver2017mastering, schrittwieser2020mastering, belousov2021reinforcement} in recent years has provided a new way for dexterous manipulation learning \cite{nagabandi2020deep, andrychowicz2020learning}. Learning complex dexterous in-hand manipulation with deep reinforcement learning has obtained impressive performance. \cite{barth2018distributed} proposed a distributed distributional deep deterministic policy gradient algorithm for dexterous manipulation tasks including catch, pick-up-and-orient, and rotate-in-hand on a simulated model of the Johns Hopkins Modular Prosthetic Limb hand with a total of 22 degrees of freedom \cite{johannes2011overview}. \cite{plappert2018multi} designed a multi-goal reinforcement learning framework for in-hand object manipulation with a Shadow Dexterous Hand \cite{tuffield2003shadow} that has five fingers with a total of 24 degrees of freedom. \cite{nagabandi2020deep} proposed a deep dynamics model for learning dexterous manipulation on a 24-DoF anthropomorphic hand, which demonstrates that a learning-based method can efficiently and effectively learn contact-rich dexterous manipulation skills. \cite{andrychowicz2020learning} used reinforcement learning to learn dexterous in-hand manipulation policies, which are trained in a simulated environment and transferred to a physical Shadow Dexterous Hand. The learning processing is conducted on a distributed RL system with a pool of 384 worker machines, each with 16 CPU cores, which can generate about 2 years of simulated experience per hour. It shows that in-hand manipulation learning with reinforcement learning in a simulator can achieve an unprecedented level of dexterity on a multi-fingered anthropomorphic hand. 
However, deep reinforcement learning for in-hand manipulation needs a large-scale dataset, which is expensive to gather. Furthermore, even the minor changes in the manipulation scenario will cause the performance sharp reduction of the source-trained model that has been trained with a large amount of data and time, as shown in Figure \ref{motivation}. 

Due to the transfer learning ability, human is able to quickly transfer learned dexterous manipulation skills to different scenarios with minimal supervision \cite{karni1998acquisition}. Transfer learning utilizes external expertise from the related well-trained task, which has been shown to effectively accelerate the learning process. Fine-tuning method is a simple transfer learning method, which fine-tunes the pre-trained model weight on a small amount of data for the target task. This method can be used for in-hand manipulation policy learning. However, the fine-tuning method just utilizes the pre-trained model weights, which is not enough for the complicated in-hand manipulation task. In the reinforcement learning procedure, the agent needs to interact with the environment to collect enough experiences for improving performance. Furthermore, at the initial training stage in the target scene, the interaction experiences include unsuccessful manipulation experiences. Fine-tuning the model with these experiences may mislead the model training to arise catastrophic forgetting, which may increase the training time and decrease the performance.
  
In order to quickly learn the complicated in-hand manipulation skills in new and similar scenarios, it is better to effectively leverage the collected trajectories and the source-trained dynamics model. Based on this consideration, we propose a new dexterous in-hand manipulation progressive transfer learning framework (PTL) for the multi-fingered anthropomorphic hand.
The proposed method adopts progressive neural networks for dexterous manipulation dynamics model transfer learning, which uses lateral connection structural with the pre-trained model. To effectively selected experiences from the trajectory replay buffer, we designed a new transfer learning samples selection method based on dynamics properties, rewards and scores of trajectories. 

Experimental results on contact-rich anthropomorphic hand manipulation tasks show that the proposed method can efficiently and effective learn the in-hand manipulation skills with a few online attempts and adjustment learning under the new scene in which the manipulation objects have been changed. The proposed method reduces the data collection time and obtains better performance than learning from scratch and fine-tuning methods.

Our main contributions can be summarized as follows:
\begin{itemize}
    \item We propose a novel dexterous in-hand manipulation progressive transfer learning framework (PTL) based on leveraging the collected trajectories and the source-trained dynamics model. The proposed framework can efficiently learn the in-hand manipulation skills with a few online attempts in the new scenes.
    \item We propose a new efficient transfer learning samples selection method  which is based on the dynamics properties, rewards and scores of the collected trajectories. Using this method, our framework can effectively select trajectories from the source domain collected trajectory replay buffer, and reduce storage requirements.
    \item We design three transfer learning tasks of dexterous in-hand manipulation to evaluate the performance. Extensive experimental results for comparison and ablation study demonstrate the effectiveness of the proposed method.
\end{itemize}


\section{Related Work}
This section provides a brief summary of several related works, including in-hand dexterous manipulation, deep transfer reinforcement learning, and experience replay.

\subsection{In-hand Dexterous Manipulation}
The goal of in-hand manipulation with a dexterous hand is to operate the objects to the desired pose by controlling fingers dynamically over time \cite{rus1999hand}. Common in-hand manipulation tasks include rotation, regrasping, and rolling through finely controlled fingers contacts \cite{ma2011dexterity, kumar2016optimal, psomopoulou2018stable, cruciani2020benchmarking}. In recent years, many model-based control methods have been explored for in-hand manipulation with underactuated hands \cite{calli2017vision, liarokapis2016learning, rojas2016gr2, caldas2022task}. The performance of these model-based methods depends on the accuracy of modeling manipulation scenarios.


In order to make robot manipulators work gracefully in practical applications, the robot manipulators must be adaptive to variations of properties of objects and the robot \cite{cui2021toward}. Machine learning is an important approach for intelligent robotics to deal with the model variation of manipulation environment \cite{kroemer2021review}. Such as, \cite{funabashi2022multi, khandate2021feasibility, sundaralingam2021hand} use machine learning methods to deal with sensing information for multi-fingered in-hand manipulation. Deep reinforcement learning has been widely used for dexterous manipulation, which has shown great potential \cite{charlesworth2021solving, ibarz2021train, gupta2021reset, nagabandi2020deep, andrychowicz2020learning}. These methods obtain good performance based on a large amount of dataset and a time-consuming training process. If the manipulation scenarios have been changed, the trained models will not work well. 


Learning a single strategy that works well in all environments is very difficult. It requires lots of training data as the methods adopted in \cite{andrychowicz2020learning, chen2021simple,huang2021generalization, chen2022system}, which generate many and diversity of training data with extensive randomized simulation environments. Due to the complexity of the in-hand manipulation task, acquiring sufficient interaction samples is not easy to implement. Therefore, in this study, we focus on the problem of how to effectively and efficiently utilize external expertise to speed up dexterous in-hand manipulation learning processing with better performance.

\subsection{Deep Transfer Reinforcement Learning}
Transfer learning in reinforcement learning transfers knowledge from some source tasks to a target task for addressing sequential decision-making problems \cite{taylor2009transfer}. The transferred knowledge can speed up the learning process and improve performance by reducing data collection and training time consumption \cite{yang2020transfer, han2022transfer}.

There are many kinds of transfer learning methods. For example, forward transfer methods that train a model on one task and transfer it to a new task, include fine-tuning method \cite{Haarnoja2017ReinforcementLW, Florensa2017StochasticNN, Kumar2020OneSI}, domain adaptation method \cite{Tzeng2016AdaptingDV, Eysenbach2021OffDynamicsRL,chen2022domain}, and randomization method \cite{tobin2017domain,peng2018sim}. These methods obtain impressive performance, while they require a large amount of dataset, computational cost, and training time. Multi-task transfer methods, which train a model on many tasks and transfer it to a new task, include policy distillation methods \cite{ Rusu2016PolicyD, Berseth2018ProgressiveRL}, and contextual policy method \cite{Gimelfarb2021ContextualPT}. Transfer learning methods based on models and policies, include transferring with learned models \cite{sun2021model, Sasso2022MultiSourceTL}, learned policy \cite{rusu2016progressive, rusu2017sim}, and successor representation \cite{Barreto2018TransferID,Abdolshah2021ANR}.


Transfer reinforcement learning for dexterous manipulation on a multi-fingered anthropomorphic hand is very challenging, which involves dealing with the high-dimensionality of manipulation dynamics. Prior works focus on sim-to-real transfer learning for dexterous manipulation \cite{andrychowicz2020learning, allshire2021transferring} with domain randomization strategies.  Different from previous works, we focus on using collected experiences and trained models for faster learning and better performance in the new domain task.

\subsection{Experience Replay}
Experience replay \cite{lin1992self} is used popularly for stabilizing learning in modern deep reinforcement learning \cite{mnih2015human, schaul2015prioritized, de2018experience, fedus2020revisiting}, which improves sample efficiency by reusing the stored uncorrelated experiences. Collecting diverse exploratory data can effectively improve the performance of the offline reinforcement learning algorithm, as proved in \cite{yarats2022don}. 

For transfer learning in reinforcement learning, experience replay is beneficial for the target task that only has a small number of interaction samples. In the transfer learning procedure, not all the experience samples in the source task are good for the target task learning, which due to the differences in dynamics characteristics between the source task and target task. In order to reduce the learning complexity, \cite{lazaric2008transfer} selects samples from the source task that are most relevant to the target task for transfer reinforcement learning. \cite{tirinzoni2018importance, tirinzoni2019transfer} estimate the importance weight of transfer samples from source task to target task, and assign their contribution to the learning process proportional to their importance weight. These methods achieve better transfer learning performance. \cite{yin2017knowledge} adopts hierarchical experience replay to transfer knowledge for deep reinforcement learning by selecting experiences from the replay memories, which can effectively accelerate learning with good performance. 

In our progressive transfer learning framework, we measure the action smoothness of the trajectories and combine them with rewards and scores of trajectories to select transfer experiences from the source task.

\begin{figure*}[htbp]
\centering
\includegraphics[width=6.5in]{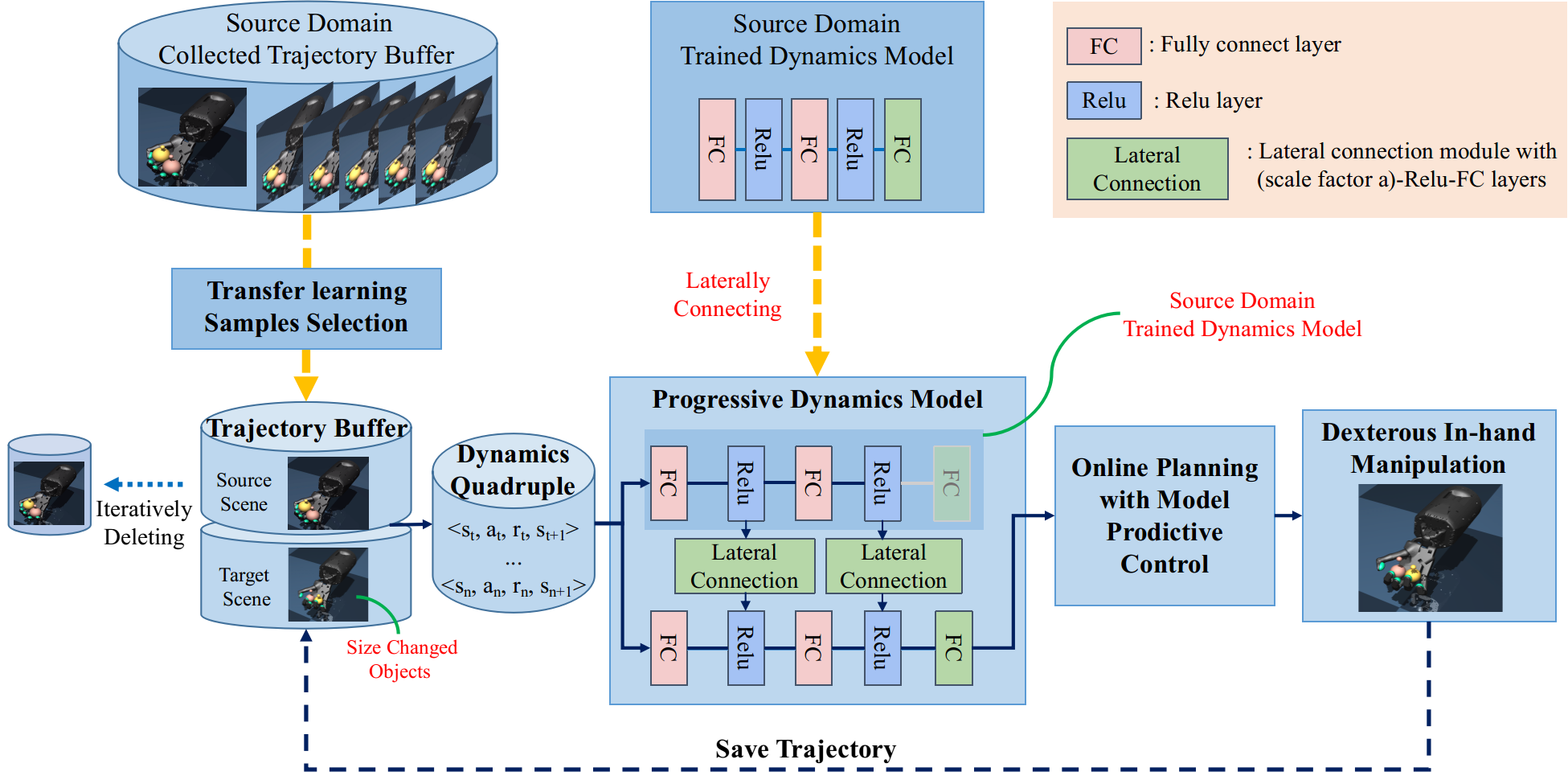}
\caption{Overview of the proposed progressive transfer learning framework for multi-fingered dexterous in-hand manipulation.}
\label{framework}
\end{figure*}

\section{Method}
In this section, we introduce the proposed progressive transfer learning method for dexterous in-hand manipulation with multi-fingered anthropomorphic hand, which includes problem statement, framework, transfer learning samples selection method, progressive dynamics model, and online planning with model predictive control.

\subsection{Problem Statement}
In this work, we focus on the problem of dexterous in-hand manipulation with faster learning and better performance based on collected experiences and trained model. We define the dynamics with a Markov decision process $\mathcal{M} = (\mathcal{S},\mathcal{A},p, R)$: state in manipulation environment $s\in\mathcal{S}$, robot action $a\in\mathcal{A}$, state transition distribution $p(s'|s, a)$, and $R$ corresponds to the reward function $r(s,a)$. The goal of manipulation policy is to select actions with maximizing the expected total rewards over a trajectory.  

\subsection{Dexterous In-Hand Manipulation Progressive Transfer Learning Framework}
Dexterous in-hand object manipulation task is complex, and learning-based methods always require large amounts of data. Our goal is to speed up transfer learning processing with better performance. Sample efficiency is an important factor that we consider. Model-based reinforcement learning approaches leverage the autonomous learning ability from data-driven methods, that makes them succeed on complex dexterous manipulation tasks. As proved in \cite{nagabandi2020deep}, for challenging dexterous in-hand manipulation, such as Baoding Balls task, the model-free methods were unable to succeed at this task. Therefore, we adopt the model-based reinforcement learning with online planning for dexterous in-hand object manipulation tasks as a baseline, which is proposed in \cite{nagabandi2020deep}. 

In order to effectively and efficiently utilize the collected experience and the learned dynamics model, we extend the baseline method with progressive neural networks for dexterous in-hand manipulation progressive transfer learning based on selected samples from source task. The overall framework (PTL) is shown in Figure \ref{framework}. The benefit of this framework is that it decouples the dynamics model learning and manipulation policy planning, allowing it to focus more on how well source tasks may transfer to target tasks in dynamics during transfer learning. As a result, the framework can simplify transfer learning while improving sample efficiency.

\subsection{Transfer Learning Samples Selection}
Learning dexterous manipulation with reinforcement learning requires a large amount of dataset, and collecting this dataset is time-consuming. In the source task, after a long time of training, it can acquire a well-trained model and collect large-scale interaction data in the manipulation environment. When switching to a new task, it is better to make full use of the collected experience data to speed up the transfer learning process and performance. Using all the source task datasets can be highly suboptimal \cite{Kalashnikov2021MTOptCM}, due to the negative transfer when transferred samples from sources being much different with the target task \cite{wiering2012reinforcement,pan2018multisource}. Moreover, using all the source samples needs more storage and computing cost. Therefore, we design a new transfer sample selection strategy that is inspired by the computational principles of movement neuroscience. 

The state of the position and velocity of the hand always changes continuously within a movement \cite{wolpert2000computational}. Based on maximizing smoothness of the hand trajectory, \cite{flash1985coordination} proposed an optimal control model of movements. Inspired by the above, we select the trajectories (rollouts) of which the states change smoothly with high total reward and final score from the source domain collected trajectory buffer. This is a prioritized experience replay strategy. Thus, we specify a priority value $PV_i$ of each trajectory for the transfer sample selection strategy, which is defined as 
\begin{equation}
\label{priority_value}
PV_i = (\lambda_1 * Reward_i + (1-\lambda_1)*MeanScore_i) * e^{-StateTD_i},
\end{equation}
where $Reward_i$ is the total reward, $MeanScore_i$ is the mean final score. $\lambda_1$ is a hyperparameter to balance the contribution of each term. We set $\lambda_1 = 0.4$ experimentally.
$StateTD_i$ is the mean state transition divergence of anthropomorphic hand in each trajectory, which is defined as 
\begin{equation}
\label{stateTD}
StateTD_i = \frac{1}{T}\sum_{t=0}^{T-1} ||s_{t+1}^{hand}-s_t^{hand}||^2,
\end{equation}
where $T$ is the total steps of each trajectory. The smaller value of $StateTD_i$ means the state changes more smoothly.

In order to select trajectories with higher priority value for target model training, 
we adopt the method proposed in \cite{daley2019reconciling} to design prioritized sampling probability $p_i$ based on priority value as follows:
\begin{equation}
\label{sample_probability}
p_i = \left\{ \begin{array}{ll}
\frac{1}{N_{rollouts}} * (1 + \rho)& \textrm{if $PV_i> median(PV)$}\\
\frac{1}{N_{rollouts}} & \textrm{if $PV_i = median(PV)$}\\
\frac{1}{N_{rollouts}}* (1 - \rho) & \textrm{if $PV_i< median(PV)$},
\end{array} \right.
\end{equation}
where $\rho \in [0,1]$ is an interpolation hyperparameter, we set $\rho = 0.1$ experimentally. $median(PV)$ is the median of all trajectory priority values $PV$. $N_{rollouts}$ is the total number of the collected trajectories (rollouts). We normalize the above value with 
\begin{equation}
\label{probability_normalize}
p_i = \frac{p_i}{\sum_i^{N_{rollouts}} p_i}.
\end{equation}

Then, we randomly sample transfer learning trajectories $D_{selected}$ from
$D_{source}$ with prioritized sampling probability $p_i$ at sample ratio $\rho_{number} = 0.1$. With these selected trajectories, we can generate dynamics quadruple $(s_t, a_t, r_t, s_{t+1})$ for progressive dynamics model learning. The algorithm of transfer learning samples selection is shown in Algorithm \ref{alg:algorithm0}.

Furthermore, we collect random trajectories $\mathcal{D}_{target-random}$ in the target scene, and get the transfer learning dataset $\mathcal{D}_{target}$ by combing selected trajectories $\mathcal{D}_{selected}$ with collected random target scene trajectories $\mathcal{D}_{target-random}$. During robot hand manipulation, we can get the target trajectory data, and store it into trajectory buffer for experience replay. Meanwhile, we delete some outdated trajectories that come from the source task scene. With this iteratively updating transferring samples strategy, we can make the dynamics model learning procedure progressively focus on the interaction data in the target scene.

\begin{algorithm}[htbp]
	\caption{Transfer Learning Samples Selection.}
	\label{alg:algorithm0}
	\KwIn{Collected Trajectories: $\mathcal{D}_{source}$ \\
     ~~~~~~~~~Sample Ratio: $\rho_{number}$\\
     ~~~~~~~~~Interpolation Hyperparameter: $\rho$}
	\KwOut{Selected Trajectories: $\mathcal{D}_{selected}$}  
	\BlankLine
	
	\For{\textnormal{each trajectory in $\mathcal{D}_{source}$}}{
       Get total reward of trajectory $i$: $Reward_i$\\
       Get mean final of trajectory $i$: $MeanScore_i$\\
       Compute robot state transition divergence: $StateTD_i$\\
       Compute priority value $PV_i$ with Equation \ref{priority_value} \\
       Compute prioritized sampling probability $p_i$ with Equation \ref{sample_probability} and \ref{probability_normalize} \\
		}
	{Sample transfer learning trajectories $\mathcal{D}_{selected}$} from $\mathcal{D}_{source}$ with prioritized sampling probability $p_i$ at sample ration $\rho_{number}$.  
\end{algorithm}

\subsection{Progressive Dynamics Model}
Inspired by \cite{nagabandi2020deep}, we use neural networks to learn an approximate model $\hat{p}_{\theta}(s'|s, a)$ for state transition distribution $p(s'|s, a)$ in manipulation dynamics, which is parameterized by $\theta$. The parameterization form of state transition distribution is based on Gaussian model with $\hat{p}_{\theta}(s'|s, a)=\mathcal{N}(\hat{f}_{\theta}(s,a), \Sigma)$, where the mean $\hat{f}_{\theta}(s,a)$ is given by the learned dynamics model with neural networks, and $\Sigma$ denotes the covariance of conditional Gaussian distribution. 

In the framework, we use progressive neural networks \cite{rusu2016progressive} to learn the dynamics model of in-hand manipulation. Different from the original networks proposed in \cite{rusu2016progressive, rusu2017sim}, some specific changes have been made to make it more suitable for in-hand manipulation transfer learning. 
First, considering the sample efficiency problem, we use progressive neural networks to learn the approximate state transition distribution $\hat{p}_{\theta}(s'|s, a)$ for manipulation dynamics, instead of learning the policy $\pi(a|s)$. 
Second, the lateral connection structure has some changes, outputs of the first and second hidden layers in the source task column are laterally connected into the first and second layers of the target task column. It is different from \cite{rusu2016progressive} where outputs of the first and second hidden layers in the source task column are laterally connected with outputs of first and second layers in the target task column. This change is intended to fully leverage the feature representation capability of the source-trained model. Through these modifications, our framework can obtain better performance than the original progressive neural networks \cite{rusu2016progressive, rusu2017sim}, which is verified in the experimental results.

Specifically, the progressive transfer learning model starts with the source task column: it is a three-layer network with hidden activations $h_{Source}^k \in \mathbb{R}^{n_{Source}^k}$, where $n_{Source}^k$ is the number of units at layer $k=\{1,2\}$. The source task column has parameters $\Theta_{Source}$ that are frozen in the transfer learning process. The target task column has the same structure with the source task column with $h_{Target}^k \in \mathbb{R}^{n_{Target}^k}$, where $n_{Target}^k$ is the number of units at layer $k=\{1,2\}$. The target task column has parameters $\Theta_{Target}$, which are initialized randomly in the transfer learning process. The first hidden layer gets input from both $h_{Target}^{0}$ (networks input) and $h_{Source}^{1}$ via lateral connections. The second hidden layer gets input from both $h_{Target}^{1}$ and $h_{Source}^{2}$ via lateral connections.

Based on the progressive dynamics model, we can get each hidden layer output with:
\begin{equation}
\label{pnn_v2}
h_{Target}^{k} = \mathcal{F}(W_{Target}^{k}h_{Target}^{k-1} + W_{lc}^{k}\mathcal{F}(\alpha_{lc}^{k}h_{Source}^{k})),
\end{equation}
where $W_{Target}^k \in \mathbb{R}^{n_{Target}^{k} \times n_{Target}^{k-1}}$ is the weight matrix of layer $k$ of the target task column. $ W_{lc}^k \in \mathbb{R}^{n_{Target}^k \times n_{Source}^k}$ denotes the lateral connection weight matrix.
$\alpha_{lc}^k$ is the learnable scalar factor in lateral connection, which is initialized randomly.
$\mathcal{F(\cdot)}$ denotes an element-wise non-linearity function $\mathcal{F}(x) = max(0,x)$. And the output of the hidden layer in source task column $h_{Source}^{k}$ is provided with:

\begin{equation}
\label{pnn_v2_h_S}
h_{Source}^{k} = \mathcal{F}(W_{Source}^{k}h_{Source}^{k-1}),
\end{equation}
where $W_{Source}^k \in \mathbb{R}^{n_{Source}^{k} \times n_{Source}^{k-1}}$ is the weight matrix of layer $k$ of the source task column.
Note that the $h_{Source}^0$ and $h_{Target}^0$ are the inputs of the networks which include the states and actions.

We learn a progressive dynamics functions $\hat{f}_\theta(s_t, a_t)$ that predicts change between $s_t$ and $s_{t+1}$. Thus, we can get the next state with : $\hat{s}_{t+1} = \hat{f}_\theta(s_t, a_t) + s_t$. And we train the progressive dynamics model with the standard mean squared error (MSE) loss as follows: 
\begin{equation}
\label{loss_func}
L = \frac{1}{|\mathcal{D}_{target}|} \sum_{(s_t, a_t, s_{t+1}) \in \mathcal{D}_{target}} \frac{1}{2}||(s_{t+1}-s_t) - \hat{f}_{\theta}(s_t, a_t)||^2,
\end{equation}
where $\mathcal{D}_{target}$ is the transfer learning dataset for the target scene. 


\subsection{Online Planning with Model Predictive Control}

Based on the learned dynamics model, we can predict the state of taking action $a$ from state $s$, and utilize this prediction to select optimal actions with online planning based on model predictive control (MPC) as the method in \cite{nagabandi2020deep}. This online planning method performs a short-horizon trajectory optimization by using the model predictions of different action sequences. It generates $N$ candidate action sequences with filtering and reward-weighted refinement, which can reduces the dimensionality of the search space, making it better scale with dimensionality. Then, it selects the optimal action sequence with the maximum reward.

\subsection{Overall Algorithm}
The overall algorithm of dexterous in-hand manipulation progressive transfer learning is shown in Algorithm \ref{alg:algorithm1}.

\begin{algorithm}[htbp]
	\caption{Progressive Transfer Learning.}
	\label{alg:algorithm1}
	\KwIn{Collected source domain trajectories: $\mathcal{D}_{source}$
          Trained source domain dynamics model: $Model_{source}=\{f_{\theta_0}^S, ..., f_{\theta_M}^S\}$}
	\KwOut{Trained dynamics model for target domain $Model_{target}=\{f_{\theta_0}^T, ..., f_{\theta_M}^T\}$}  
	\BlankLine
	Initialize the source column network in progressive dynamics model with the trained dynamics model weights $Model_{source}$ 
	
	Initialize the target column and lateral connection network in progressive dynamics model randomly
	
	  Get the selected trajectories $\mathcal{D}_{selected}$ from $\mathcal{D}_{source}$ with Algorithm \ref{alg:algorithm0}
	
	Collect random trajectories $\mathcal{D}_{target-random}$ in target scene
	
	Get the transfer learning dataset $\mathcal{D}_{target}$ by combing samples selection $\mathcal{D}_{selected}$ with $\mathcal{D}_{target-random}$
	\BlankLine
	
	\For{\textnormal{iter in range($I$)}}{  
		\For{\textnormal{rollout in range($RN$)}}{
			Reset environment $s_0$
			
			\For{\textnormal{t in range($T$)}}{
			 $a_t$  $\leftarrow$ \textnormal{Online Planning with MPC}
			
			$s_{t+1}$, $r_t$  $\leftarrow$ \textnormal{take~ action~ $a_t$}
			
			$\mathcal{D}_{target}$ $\leftarrow$ $(s_t, a_t, r_t, s_{t+1})$
			}
		}
		\BlankLine
		Update $\mathcal{D}_{target}$ by deleting outdated data from $\mathcal{D}_{selected}$
		
		Train $Model_{target}$ with $\mathcal{D}_{target}$ 
		}	
\end{algorithm}

\section{Experiments}

In this section, we will introduce the transfer learning tasks, experimental setting, measurement methods, comparison results, and ablation study.

\subsection{Transfer Learning Tasks}
\begin{figure}[!t]
\centering
\includegraphics[width=3.3in]{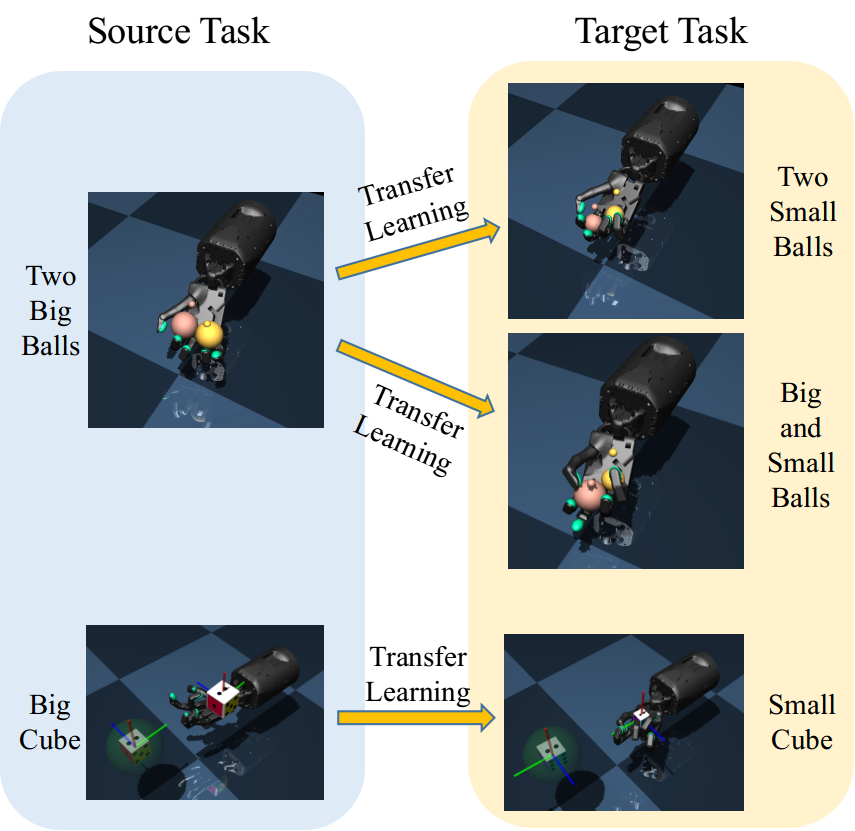}
\caption{Transfer learning task examples for dexterous in-hand manipulation.}
\label{task_example}
\end{figure}

We evaluate our method on some challenging tasks of dexterous in-hand manipulation that include Baoding Balls task and Cube Reorientation task, as shown in Figure \ref{task_example}. 

The Baoding Balls task aims to rotate two balls around the palm with a multi-fingered anthropomorphic hand. It is a contact-rich task that includes not only contacts between balls and hand, but also contacts between two balls. This leads to discontinuous dynamics and difficulty to keep two balls in the hand during rotating two balls. We design a transfer learning problem by rotating different size balls, which is an inter-task transfer learning problem \cite{yang2020transfer}. The source task is to rotate two big balls, and the target task is to rotate two small balls. They have the same state-action space. However, their transition dynamics are different due to different ball sizes. Thus, the same action produces different influences on the same states in the source and target tasks. As shown in Figure \ref{quality_picture}, the policies of rotating different size balls are different. When rotating big size balls, it always uses the little finger to push the big ball up in the source task, while it uses the ring finger to push the small ball up and the little finger keeps the ball from dropping in the target task. Even though these two tasks are very similar, a well-trained model in the source task is not work in the target task, which is as shown in Figure \ref{quality_picture}. 

To further evaluate the transfer learning performance, we design a new Baoding Balls task that aims to rotate one big ball and one small ball simultaneously. Meanwhile, we design a transfer learning problem that transfers rotating two balls with the same size to different sizes, as shown in Figure \ref{task_example}.

Furthermore, we also evaluate the transfer learning performance on Cube Reorientation task with different size cubes, as shown in Figure \ref{task_example}. Cube Reorientation task aims to reorient a cube to a goal pose in hand, which needs to avoid dropping the cube from the palm.

\subsection{Experimental Setting}
In our experiments, the manipulation environment setting is the same as in the baseline \cite{nagabandi2020deep}. In Baoding Balls task, the dimension of the state space is $40$, and the dimension of the action space is $24$. The frequency of action control is 10 per second. The reward function is defined as:
\begin{equation}
\label{reward_func_baoding}
\begin{split}
 r=-5||object_{xyz}-target_{xyz}||-500 * \mathbbm{1} (isdrop) \\
 - 500 * \mathbbm{1} (toohigh),   
\end{split}
\end{equation}
where $object_{xyz}$ and $target_{xyz}$ denote the coordinate of the balls at current and target respectively. The term $-500 * \mathbbm{1} (isdrop)$ means to penalize for the ball dropping from the hand, and the term $- 500 * \mathbbm{1} (toohigh)$ means to penalize wrist angle for lifting up too much. Its score is defined as:
\begin{equation}
\label{score_func_baoding}
Score = -||object_{xyz}-target_{xyz}||.
\end{equation}

In Cube Reorientation task, the dimension of the state space is $36$, and the dimension of the action space is $24$. The reward function is defined as:
\begin{equation}
\label{reward_func_cube}
r = -7||cube_{rpy}-target_{rpy}||-100 * \mathbbm{1} (isdrop),
\end{equation}
where $cube_{rpy}$ and $target_{rpy}$ denote the rotational orientation of the cube at current and target respectively.
For Cube Reorientation task, its score is defined as:
\begin{equation}
\label{score_func_cube}
Score = -||cube_{rpy}-target_{rpy}||.
\end{equation}

The hyperparameters for Baoding Balls and Cube Reorientation tasks are set the same as in \cite{nagabandi2020deep}, including $RN=30, T=100, H=7, N=700 ~\text{(Balling Balls)}, N=1000 ~\text{(Cube Reorientation)}$.


\subsection{Measurement Methods}
In order to measure the performance, we evaluate the total rewards and mean final scores that are averaged over five random seeds. Meanwhile, we measure success rates ($SR$) with rollout total reward ($Reward_i$) and mean final score ($Score_i$) above some thresholds (reward threshold: $RT$, score threshold: $ST$), which are defined as follows:
\begin{equation}
\label{succeed_rate_reward}
SR_{Reward} = \frac{\sum_{i=1}^{N_{rollouts}} \mathbbm{1} (Reward_i >=  RT)  }{N_{rollouts}},
\end{equation}

\begin{equation}
\label{succeed_rate_score}
SR_{Score} = \frac{\sum_{i=1}^{N_{rollouts}} \mathbbm{1} ( Score_i >= ST)  }{N_{rollouts}},
\end{equation}
For Baoding Balls task, $RT=-20,ST=-0.02$. For Cube Reorientation task, $RT=-250,ST=-0.5$.

\begin{table*}[htbp]
\caption{Transfer learning performance comparisons on Baoding Balls task with the same size change of the two balls. Succeed Rate with Reward threshold (-0.20). Succeed Rate with Score threshold (-0.02). \label{tab:table0}}
\centering
\begin{threeparttable}
\begin{tabular}{|c|c|c|c|c|c|c|}
\hline
\multicolumn{1}{|c|}{\multirow{2}{1.2cm}{Method}} & \multicolumn{1}{c|}{\multirow{2}{1.8cm}{Reward}}  & \multicolumn{1}{c|}{\multirow{2}{1.8cm}{Score}} & \multicolumn{1}{c|}{\multirow{2}{2cm}{Succeed Rate with Reward (\%)}} &\multicolumn{1}{c|}{\multirow{2}{1.9cm}{Succeed Rate with Score (\%)}} & \multicolumn{1}{c|}{\multirow{2}{1.2cm}{Training Time (H)}} & \multicolumn{1}{c|}{\multirow{2}{2.0cm}{Number of Datapoints }} \\
&&&&&&\\
\hline
Source task learning &-28.920   &-0.02  &95.0 &80.0 &110 &1,540,080\\
\hline
Learning from scratch & $-124.883 \pm 65.948$  & $-0.062 \pm 0.002$ &76.7 &3.0 &31 &914,034\\
\hline
Fine-tuning from scratch & $-142.851 \pm 30.345$ & $-0.064 \pm 0.005$ &73.3 &7.0 &\textbf{2.7} &285,046\\
\hline
Fine-tuning &$-29.553 \pm 5.078$  & $-0.023 \pm 0.001$ &95.0 &61.5 &3.5 &\textbf{177,664}  \\
\hline
PNN\cite{rusu2016progressive} &$-9.800 \pm 6.112$ &$-0.015 \pm 0.005$ &98.8 &82.0 &4.4 &389,001 \\
\hline
PNN-V1\cite{rusu2017sim} &$-15.822 \pm 10.152$ &$-0.013 \pm 0.003$ &97.6 &86.8  &4.1 &395,945\\
\hline
PTL (Our) & $\bm{-7.456 \pm 2.035}$ & $\bm{-0.010 \pm 0.001}$ &\textbf{99.2}   &\textbf{92.8}  &4.7 &189,797\\
\hline
\end{tabular}
\begin{tablenotes}
\footnotesize
 \item[1] Except Learning from scratch is the result of training with 300 iterations, source task learning is the result of training with 560 iterations, others are the results of training with 50 iterations.
 \item[2] Except noting "from scratch" is training from scratch, others are trained with selected samples.
 \item[3] Except source task learning is trained with one random seed, other results are averaged over five random seeds.
\end{tablenotes}
\end{threeparttable}
\end{table*}

\begin{figure*}[htbp]
\centering
\subfloat[]{\includegraphics[width=0.5\textwidth]{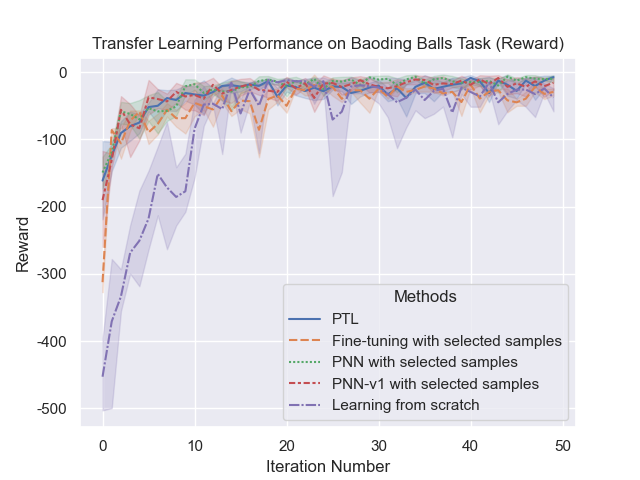}%
\label{fig_SRR}}
\hfil
\subfloat[]{\includegraphics[width=0.5\textwidth]{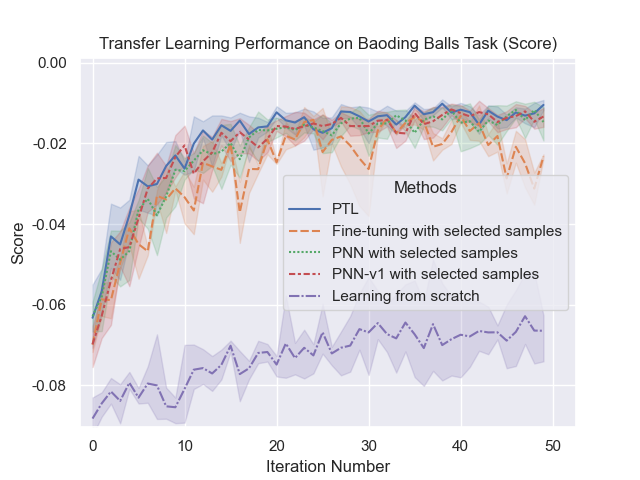}%
\label{fig_SRS}}
\caption{Transfer learning performance of the training process on Baoding Balls task with the same size change scene. (a) Rewards for model performance after each iteration of training. (b) Scores for model performance after each iteration of training.}
\label{fig_reward_score}
\end{figure*}

\begin{figure*}[htbp]
\centering
\includegraphics[width=6.8in]{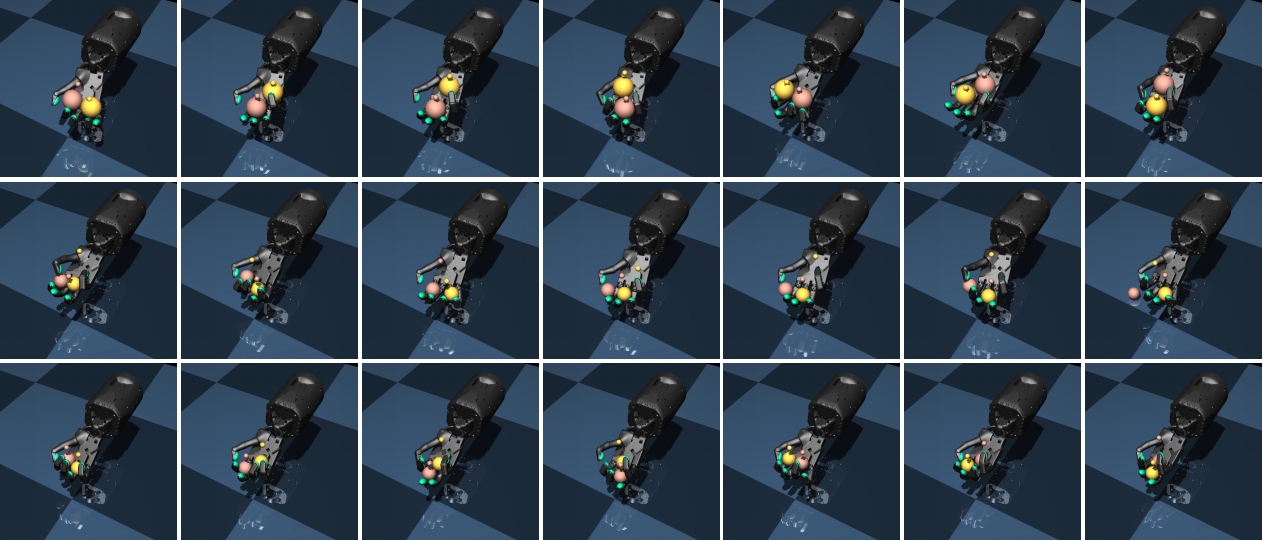}
\caption{Quality results on Baoding Balls task with two small balls. a) First row shows the results of the source-trained model on the source task. b) Second row shows the results of the source-trained model on the target task. c) Third row shows the results of the PTL model on the target task.}
\label{quality_picture}
\end{figure*}

\subsection{Comparison Results}
For the transfer learning problem on Baoding Balls task with the same size change of the two balls, we compare the proposed method with learning from scratch, fine-tuning from scratch, fine-tuning with selected samples, PNN\cite{rusu2016progressive} and PNN-V1\cite{rusu2017sim}. Note that PNN\cite{rusu2016progressive} and PNN-V1\cite{rusu2017sim} are model-free approaches that are unable to succeed at Baoding Balls task as proved in \cite{nagabandi2020deep}. Therefore, we modify them into our framework, which adopts the progressive neural networks for transition dynamics learning instead of learning the policy directly. As shown in Table \ref{tab:table0}, PTL can reduce the training time with better performance. Compare with methods that learn or fine-tune from scratch, training with experience replay not only saves time, but also obtains better performance. Learning or fine-tuning from scratch may get stuck in pursuit of high reward, while the scores have not been improved, which means it is unable to complete the task. It shows that the successful experiences on the source task can guide the training with better performance on the target task. Figure \ref{fig_reward_score} shows that the proposed method can obtain more robust training performance than fine-tuning method. The manipulation examples are shown in Figure \ref{quality_picture}. We can find that the PTL model is able to effectively coordinate the high degree-of-freedom fingers, which can manipulate the two balls successfully.

\begin{figure*}[htbp]
\centering
\includegraphics[width=6.8in]{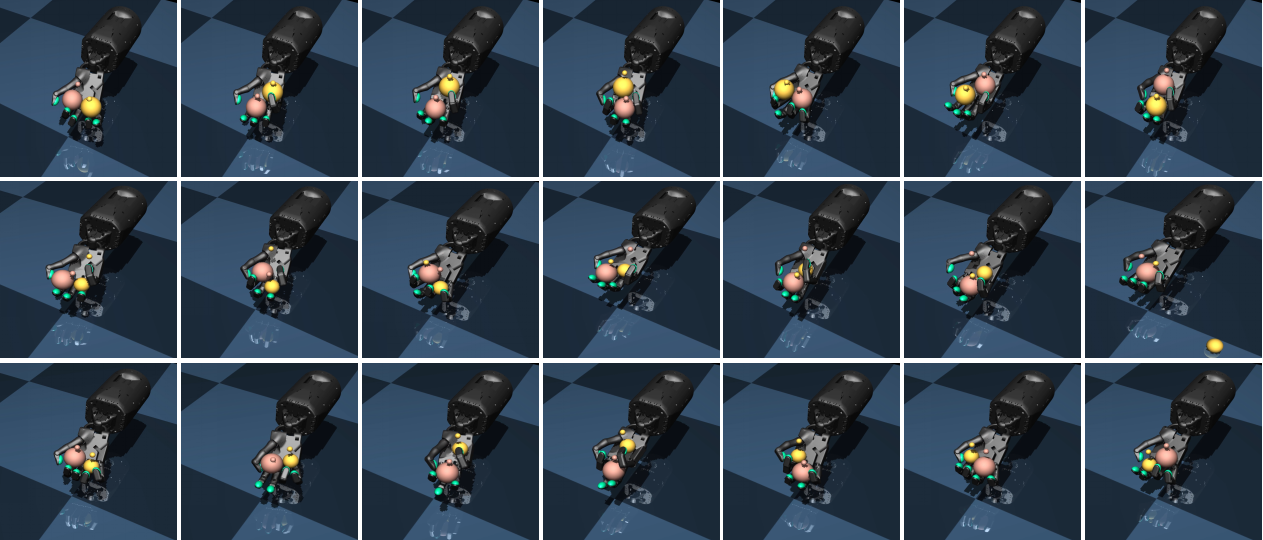}
\caption{Quality results on Baoding Balls task with big and small balls. a) First row shows the results of the trained model on the source task. b) Second row shows the results of the source-trained model on the target task. c) Third row shows the results of the PTL model on the target task.}
\label{quality_picture2}
\end{figure*}

For the transfer learning problem on Baoding Balls task with big and small balls, we compare our method with learning from scratch and fine-tuning methods. For this more challenging problem, our method can also achieve better performance, as shown in Table \ref{tab:table1}. From the manipulation examples shown in Figure \ref{quality_picture2}, we can find that the source-trained model cannot predict the state transition accuracy, which makes it unable to select optimal actions with online planning method. Based on our PTL framework, our method can learn the dynamics of the new manipulation scenario more accurately. The more precise dynamics prediction enables the multi-fingered anthropomorphic hand to better complete the dexterous in-hand manipulation with online planning.

\begin{table*}[htpb]
\caption{Transfer learning performance comparisons on Baoding Balls task with big and small balls. Succeed Rate with Reward threshold (-0.20). Succeed Rate with Score threshold (-0.02). \label{tab:table1}}
\centering
\begin{threeparttable}
\begin{tabular}{|c|c|c|c|c|c|c|}
\hline
\multicolumn{1}{|c|}{\multirow{2}{1.2cm}{Method}} & \multicolumn{1}{c|}{\multirow{2}{1.8cm}{Reward}}  & \multicolumn{1}{c|}{\multirow{2}{1.8cm}{Score}} & \multicolumn{1}{c|}{\multirow{2}{2cm}{Succeed Rate with Reward (\%)}} &\multicolumn{1}{c|}{\multirow{2}{1.8cm}{Succeed Rate with Score (\%)}} & \multicolumn{1}{c|}{\multirow{2}{1.2cm}{Training Time (H)}} & \multicolumn{1}{c|}{\multirow{2}{1.6cm}{Number of Datapoints }} \\
&&&&&&\\
\hline
Learning from scratch &-117.527  & -0.064 &78.0 &1.0 &10 &747,097\\
\hline
Fine-tuning &$-61.926 \pm 41.780$  & $-0.027 \pm 0.006$ &88.6 &64.8 &3.1 &\textbf{735,030}  \\
\hline
PTL (Our) & $\bm{-31.160 \pm 16.227}$ & $\bm{-0.024 \pm 0.005}$ &\textbf{94.6}   &\textbf{65.8}  &\textbf{3.0} &751,592\\
\hline
\end{tabular}
\begin{tablenotes}
\footnotesize
\item[1] Except Learning from scratch is the result of training with 300 iterations, others are the results of training with 100 iterations.
\item[2] Fine-tuning and PTL are trained with selected samples.
\item[3] Except learning from scratch is trained with one random seed, other results are averaged over five random seeds.
\end{tablenotes}
\end{threeparttable}
\end{table*}

\begin{table*}[htbp]
\caption{Transfer learning performance comparisons on Cube Reorientation task with different size cubes. Succeed Rate with Reward threshold (-250). Succeed Rate with Score threshold (-0.5). \label{tab:table2}}
\centering
\begin{threeparttable}
\begin{tabular}{|c|c|c|c|c|c|c|}
\hline
\multicolumn{1}{|c|}{\multirow{2}{1.2cm}{Method}} & \multicolumn{1}{c|}{\multirow{2}{1.8cm}{Reward}}  & \multicolumn{1}{c|}{\multirow{2}{1.8cm}{Score}} & \multicolumn{1}{c|}{\multirow{2}{2cm}{Succeed Rate with Reward (\%)}} &\multicolumn{1}{c|}{\multirow{2}{1.9cm}{Succeed Rate with Score (\%)}} & \multicolumn{1}{c|}{\multirow{2}{1.2cm}{Training Time (H)}} & \multicolumn{1}{c|}{\multirow{2}{1.6cm}{Number of Datapoints}} \\
&&&&&&\\
\hline
Source task learning & -405.013  & -0.548 &66 &60 &29 &1,016,654\\
\hline
Learning from scratch 300 & $-315.633 \pm 47.151$  & $-0.390 \pm 0.034$ & 68.6 &69.0 &\textbf{6.5} &623,944\\
\hline
Learning from scratch 600 & $-290.449 \pm 29.549$  & $-0.317 \pm 0.029$ & 76.2 &76.0 &19.3 &1,266,063\\
\hline
Fine-tuning &$-327.020 \pm 45.184$  & $-0.388 \pm 0.049$ &71.6 &69.4 &9 &935,888  \\
\hline
PTL learning from scratch &$-281.960 \pm 31.900$  & $\bm{-0.291 \pm 0.036}$ &78.0 &\textbf{78.0} &19.3 &\textbf{613,799}  \\
\hline
PTL (Our) & $\bm{-273.641 \pm 30.018}$ &$-0.301 \pm 0.060$ &\textbf{78.2}   &77.8  &12.5 &1,286,745\\
\hline
\end{tabular}
\begin{tablenotes}
\footnotesize
\item[1] Except Learning from scratch 300 \& 600 are the results of training with 300 and 600 iterations, source task learning is the result of training with 590 iterations, others are the results of training with 100 iterations.
\item[2] Fine-tuning and PTL are trained with selected samples.
\item[3] Except source task learning is trained with one random seed, other results are averaged over five random seeds.
\end{tablenotes}
\end{threeparttable}
\end{table*}

\begin{figure*}[htbp]
\centering
\includegraphics[width=6.8in]{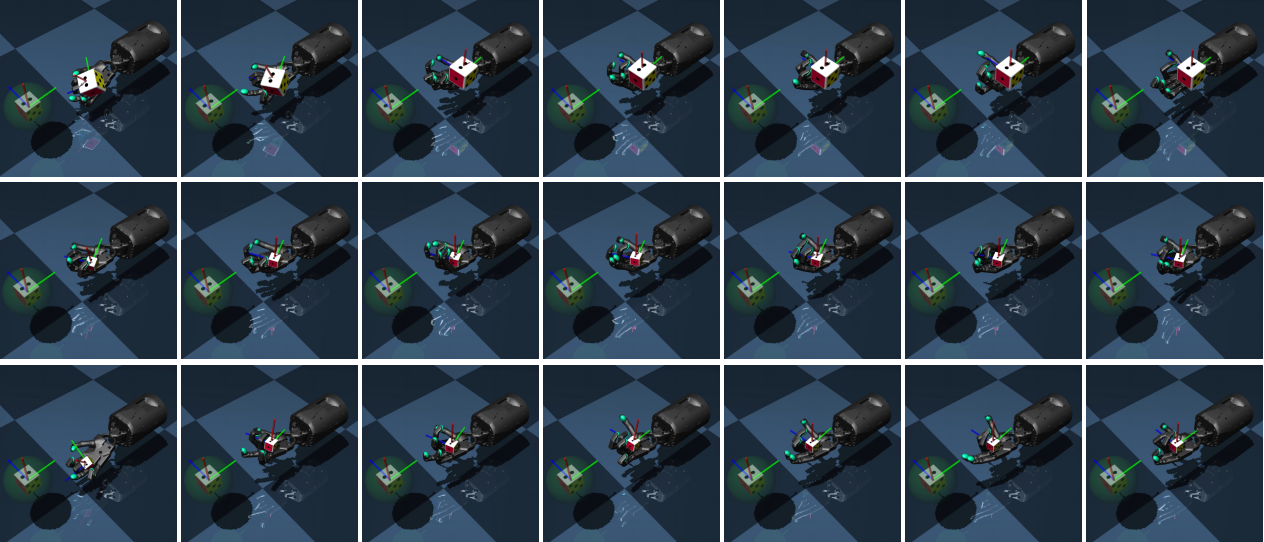}
\caption{Quality results on Cube Reorientation task with different size cubes. a) First row shows the results of the trained model on the source task. b) Second row shows the results of the source-trained model on the target task. c) Third row shows the results of the PTL model on the target task.}
\label{quality_picture3}
\end{figure*}

For the transfer learning problem on Cube Reorientation task with different size cubes, we compare our method (with or without selected samples) with learning from scratch and fine-tuning methods. For this more challenging problem, progressive transfer learning with selected samples from the source task, can reduce the time of data collection and model training, as shown in Table \ref{tab:table2}. After progressive transfer learning, the PTL model can achieve more accurate reorientation than the source-trained model, which is shown in Figure \ref{quality_picture3}.

\begin{table*}[t]
\caption{Ablation experiments about training sample selection methods. Scratch: Learning from scratch without experience data. Uniform: Learning with random uniform selected samples from source task. Select: Learning with selected samples by our proposed method from source task. Succeed Rate with Reward threshold (-0.20). Succeed Rate with Score threshold (-0.02).\label{tab:table3}}
\centering
\begin{threeparttable}
\begin{tabular}{|c|ccc|c|c|c|c|c|c|}
\hline
\multirow{2}{*}{Method} & \multicolumn{3}{|c|}{Training Examples}&  \multicolumn{1}{c|}{\multirow{2}{1.75cm}{Reward}}  &  \multicolumn{1}{c|}{\multirow{2}{1.55cm}{Score}} & \multirow{2}{2cm}{Succeed Rate with Reward (\%)} &\multirow{2}{1.75cm}{Succeed Rate with Score (\%)} & \multicolumn{1}{c|}{\multirow{2}{1.1cm}{Training Time (H)}}  & \multicolumn{1}{c|}{\multirow{2}{1.3cm}{Number of Datapoints}} \\ 
\cline{2-4}
       &Scratch &Uniform &Select  &&&&&&\\
\hline
FT  & $\surd$ & & & $-142.851 \pm 30.345$ & $-0.064 \pm 0.005$ &73.3 &7.0 &\textbf{2.7} &285, 046\\
\hline
FT  & &$\surd$ & & $-36.260 \pm 18.789$  & $-0.019 \pm 0.003$ &93.6 &73.8 &3.5 &\textbf{175, 366}\\
\hline
FT & & &$\surd$ &$-29.553 \pm 5.078$  & $-0.023 \pm 0.001$ &95.0 &61.5 &3.5 &177, 664  \\
\hline
PTL & $\surd$ & & &$-17.819 \pm 4.456$ &$-0.019 \pm 0.006$ &97.25 &73.5 &20.3 &624, 110  \\
\hline
PTL & &$\surd$ &  &$-26.043 \pm 17.233$ &$-0.014 \pm 0.003$ &95.0 &85.0  &4.3 &185, 982\\
\hline
PTL & & &$\surd$  & $\bm{-7.456 \pm 2.035}$ & $\bm{-0.010 \pm 0.001}$ &\textbf{99.2}   &\textbf{92.8}  &4.7 &189, 797\\
\hline
\end{tabular}
\begin{tablenotes}
\footnotesize
    \item[1] Except PTL scratch is the result of training with 200 epochs, others are the results of training with 50 epochs.
    \item[2] All the results are averaged over five random seeds.
\end{tablenotes}
\end{threeparttable}
\end{table*}

\subsection{Ablation Study}
In order to evaluate the effectiveness of each proposed component, we conduct some ablation experiments on Baoding Balls task with the same size change of the two balls. The ablation experiments include:
Scratch---learning from scratch without experience; Uniform---learning with random uniform samples from source task; Select---learning with samples selected by the proposed method. 

From Table \ref{tab:table3}, we can find that progressive transfer learning is able to achieve better performance than fine-tuning methods, and reduce the training time. Progressive transfer learning can also avoid getting stuck in a low score with high reward, which can complete the manipulation task better. It  can also find that using the experiences sampled from the source task can effectively reduce the data collection time and guide the model training toward better performance.

\section{Conclusion}
We have proposed a new dexterous in-hand manipulation progressive transfer learning framework for the multi-fingered anthropomorphic hand. The proposed framework exploits transfer learning to speed up learning for the new scenarios with a progressive dynamics model and experience replay based on transferring sample selection. Experiments on contact-rich anthropomorphic hand manipulation tasks demonstrate that the proposed method can efficiently and effectively learn the in-hand manipulation skills with some online attempts and adjustment learning in the new scenarios. 

In our current work, the transfer learning experiments are only conducted on manipulation object size changed scene. In the future work, we will focus on more challenging transfer learning tasks with more domain differences scenarios. Meanwhile, we will apply the proposed method to the real multi-fingered anthropomorphic hand platform.

\bibliographystyle{IEEEtran}
\bibliography{IEEEabrv,./IEEEexample.bib}

\begin{thebibliography}{10}
\providecommand{\url}[1]{#1}
\csname url@samestyle\endcsname
\providecommand{\newblock}{\relax}
\providecommand{\bibinfo}[2]{#2}
\providecommand{\BIBentrySTDinterwordspacing}{\spaceskip=0pt\relax}
\providecommand{\BIBentryALTinterwordstretchfactor}{4}
\providecommand{\BIBentryALTinterwordspacing}{\spaceskip=\fontdimen2\font plus
\BIBentryALTinterwordstretchfactor\fontdimen3\font minus
  \fontdimen4\font\relax}
\providecommand{\BIBforeignlanguage}[2]{{%
\expandafter\ifx\csname l@#1\endcsname\relax
\typeout{** WARNING: IEEEtran.bst: No hyphenation pattern has been}%
\typeout{** loaded for the language `#1'. Using the pattern for}%
\typeout{** the default language instead.}%
\else
\language=\csname l@#1\endcsname
\fi
#2}}
\providecommand{\BIBdecl}{\relax}
\BIBdecl

\bibitem{okamura2000overview}
A.~M. Okamura, N.~Smaby, and M.~R. Cutkosky, ``An overview of dexterous
  manipulation,'' in \emph{Proceedings 2000 ICRA. Millennium Conference. IEEE
  International Conference on Robotics and Automation. Symposia Proceedings
  (Cat. No. 00CH37065)}, vol.~1.\hskip 1em plus 0.5em minus 0.4em\relax IEEE,
  2000, pp. 255--262.

\bibitem{kumar2014real}
V.~Kumar, Y.~Tassa, T.~Erez, and E.~Todorov, ``Real-time behaviour synthesis
  for dynamic hand-manipulation,'' in \emph{2014 IEEE International Conference
  on Robotics and Automation (ICRA)}.\hskip 1em plus 0.5em minus 0.4em\relax
  IEEE, 2014, pp. 6808--6815.

\bibitem{bicchi2000hands}
A.~Bicchi, ``Hands for dexterous manipulation and robust grasping: A difficult
  road toward simplicity,'' \emph{IEEE Transactions on robotics and
  automation}, vol.~16, no.~6, pp. 652--662, 2000.

\bibitem{ma2011dexterity}
R.~R. Ma and A.~M. Dollar, ``On dexterity and dexterous manipulation,'' in
  \emph{2011 15th International Conference on Advanced Robotics (ICAR)}.\hskip
  1em plus 0.5em minus 0.4em\relax IEEE, 2011, pp. 1--7.

\bibitem{huang2000mechanics}
W.~H. Huang and M.~T. Mason, ``Mechanics, planning, and control for tapping,''
  \emph{The International Journal of Robotics Research}, vol.~19, no.~10, pp.
  883--894, 2000.

\bibitem{erdmann1998exploration}
M.~Erdmann, ``An exploration of nonprehensile two-palm manipulation,''
  \emph{The International Journal of Robotics Research}, vol.~17, no.~5, pp.
  485--503, 1998.

\bibitem{bai2014dexterous}
Y.~Bai and C.~K. Liu, ``Dexterous manipulation using both palm and fingers,''
  in \emph{2014 IEEE International Conference on Robotics and Automation
  (ICRA)}.\hskip 1em plus 0.5em minus 0.4em\relax IEEE, 2014, pp. 1560--1565.

\bibitem{tahara2010dynamic}
K.~Tahara, S.~Arimoto, and M.~Yoshida, ``Dynamic object manipulation using a
  virtual frame by a triple soft-fingered robotic hand,'' in \emph{2010 IEEE
  International Conference on Robotics and Automation}.\hskip 1em plus 0.5em
  minus 0.4em\relax IEEE, 2010, pp. 4322--4327.

\bibitem{li2014learning}
M.~Li, H.~Yin, K.~Tahara, and A.~Billard, ``Learning object-level impedance
  control for robust grasping and dexterous manipulation,'' in \emph{2014 IEEE
  International Conference on Robotics and Automation (ICRA)}.\hskip 1em plus
  0.5em minus 0.4em\relax IEEE, 2014, pp. 6784--6791.

\bibitem{mnih2015human}
V.~Mnih, K.~Kavukcuoglu, D.~Silver, A.~A. Rusu, J.~Veness, M.~G. Bellemare,
  A.~Graves, M.~Riedmiller, A.~K. Fidjeland, G.~Ostrovski \emph{et~al.},
  ``Human-level control through deep reinforcement learning,'' \emph{nature},
  vol. 518, no. 7540, pp. 529--533, 2015.

\bibitem{silver2017mastering}
D.~Silver, J.~Schrittwieser, K.~Simonyan, I.~Antonoglou, A.~Huang, A.~Guez,
  T.~Hubert, L.~Baker, M.~Lai, A.~Bolton \emph{et~al.}, ``Mastering the game of
  go without human knowledge,'' \emph{nature}, vol. 550, no. 7676, pp.
  354--359, 2017.

\bibitem{schrittwieser2020mastering}
J.~Schrittwieser, I.~Antonoglou, T.~Hubert, K.~Simonyan, L.~Sifre, S.~Schmitt,
  A.~Guez, E.~Lockhart, D.~Hassabis, T.~Graepel \emph{et~al.}, ``Mastering
  atari, go, chess and shogi by planning with a learned model,'' \emph{Nature},
  vol. 588, no. 7839, pp. 604--609, 2020.

\bibitem{belousov2021reinforcement}
B.~Belousov, H.~Abdulsamad, P.~Klink, S.~Parisi, and J.~Peters,
  \emph{Reinforcement Learning Algorithms: Analysis and Applications}.\hskip
  1em plus 0.5em minus 0.4em\relax Springer, 2021.

\bibitem{nagabandi2020deep}
A.~Nagabandi, K.~Konolige, S.~Levine, and V.~Kumar, ``Deep dynamics models for
  learning dexterous manipulation,'' in \emph{Conference on Robot
  Learning}.\hskip 1em plus 0.5em minus 0.4em\relax PMLR, 2020, pp. 1101--1112.

\bibitem{andrychowicz2020learning}
O.~M. Andrychowicz, B.~Baker, M.~Chociej, R.~Jozefowicz, B.~McGrew,
  J.~Pachocki, A.~Petron, M.~Plappert, G.~Powell, A.~Ray \emph{et~al.},
  ``Learning dexterous in-hand manipulation,'' \emph{The International Journal
  of Robotics Research}, vol.~39, no.~1, pp. 3--20, 2020.

\bibitem{barth2018distributed}
G.~Barth-Maron, M.~W. Hoffman, D.~Budden, W.~Dabney, D.~Horgan, D.~Tb,
  A.~Muldal, N.~Heess, and T.~Lillicrap, ``Distributed distributional
  deterministic policy gradients,'' \emph{arXiv preprint arXiv:1804.08617},
  2018.

\bibitem{johannes2011overview}
M.~S. Johannes, J.~D. Bigelow, J.~M. Burck, S.~D. Harshbarger, M.~V. Kozlowski,
  and T.~Van~Doren, ``An overview of the developmental process for the modular
  prosthetic limb,'' \emph{Johns Hopkins APL Technical Digest}, vol.~30, no.~3,
  pp. 207--216, 2011.

\bibitem{plappert2018multi}
M.~Plappert, M.~Andrychowicz, A.~Ray, B.~McGrew, B.~Baker, G.~Powell,
  J.~Schneider, J.~Tobin, M.~Chociej, P.~Welinder \emph{et~al.}, ``Multi-goal
  reinforcement learning: Challenging robotics environments and request for
  research,'' \emph{arXiv preprint arXiv:1802.09464}, 2018.

\bibitem{tuffield2003shadow}
P.~Tuffield and H.~Elias, ``The shadow robot mimics human actions,''
  \emph{Industrial Robot: An International Journal}, 2003.

\bibitem{karni1998acquisition}
A.~Karni, G.~Meyer, C.~Rey-Hipolito, P.~Jezzard, M.~M. Adams, R.~Turner, and
  L.~G. Ungerleider, ``The acquisition of skilled motor performance: fast and
  slow experience-driven changes in primary motor cortex,'' \emph{Proceedings
  of the National Academy of Sciences}, vol.~95, no.~3, pp. 861--868, 1998.

\bibitem{rus1999hand}
D.~Rus, ``In-hand dexterous manipulation of piecewise-smooth 3-d objects,''
  \emph{The International Journal of Robotics Research}, vol.~18, no.~4, pp.
  355--381, 1999.

\bibitem{kumar2016optimal}
V.~Kumar, E.~Todorov, and S.~Levine, ``Optimal control with learned local
  models: Application to dexterous manipulation,'' in \emph{2016 IEEE
  International Conference on Robotics and Automation (ICRA)}.\hskip 1em plus
  0.5em minus 0.4em\relax IEEE, 2016, pp. 378--383.

\bibitem{psomopoulou2018stable}
E.~Psomopoulou, D.~Karashima, Z.~Doulgeri, and K.~Tahara, ``Stable pinching by
  controlling finger relative orientation of robotic fingers with rolling soft
  tips,'' \emph{Robotica}, vol.~36, no.~2, pp. 204--224, 2018.

\bibitem{cruciani2020benchmarking}
S.~Cruciani, B.~Sundaralingam, K.~Hang, V.~Kumar, T.~Hermans, and D.~Kragic,
  ``Benchmarking in-hand manipulation,'' \emph{IEEE Robotics and Automation
  Letters}, vol.~5, no.~2, pp. 588--595, 2020.

\bibitem{calli2017vision}
B.~Calli and A.~M. Dollar, ``Vision-based model predictive control for
  within-hand precision manipulation with underactuated grippers,'' in
  \emph{2017 IEEE International Conference on Robotics and Automation
  (ICRA)}.\hskip 1em plus 0.5em minus 0.4em\relax IEEE, 2017, pp. 2839--2845.

\bibitem{liarokapis2016learning}
M.~V. Liarokapis and A.~M. Dollar, ``Learning task-specific models for
  dexterous, in-hand manipulation with simple, adaptive robot hands,'' in
  \emph{2016 IEEE/RSJ International Conference on Intelligent Robots and
  Systems (IROS)}.\hskip 1em plus 0.5em minus 0.4em\relax IEEE, 2016, pp.
  2534--2541.

\bibitem{rojas2016gr2}
N.~Rojas, R.~R. Ma, and A.~M. Dollar, ``The gr2 gripper: An underactuated hand
  for open-loop in-hand planar manipulation,'' \emph{IEEE Transactions on
  Robotics}, vol.~32, no.~3, pp. 763--770, 2016.

\bibitem{caldas2022task}
A.~Caldas, M.~Grossard, M.~Makarov, and P.~Rodriguez-Ayerbe, ``Task-level
  dexterous manipulation with multifingered hand under modeling
  uncertainties,'' \emph{Journal of Dynamic Systems, Measurement, and Control},
  vol. 144, no.~3, 2022.

\bibitem{cui2021toward}
J.~Cui and J.~Trinkle, ``Toward next-generation learned robot manipulation,''
  \emph{Science robotics}, vol.~6, no.~54, p. eabd9461, 2021.

\bibitem{kroemer2021review}
O.~Kroemer, S.~Niekum, and G.~D. Konidaris, ``A review of robot learning for
  manipulation: Challenges, representations, and algorithms,'' \emph{Journal of
  machine learning research}, vol.~22, no.~30, 2021.

\bibitem{funabashi2022multi}
S.~Funabashi, T.~Isobe, F.~Hongyi, A.~Hiramoto, A.~Schmitz, S.~Sugano, and
  T.~Ogata, ``Multi-fingered in-hand manipulation with various object
  properties using graph convolutional networks and distributed tactile
  sensors,'' \emph{IEEE Robotics and Automation Letters}, 2022.

\bibitem{khandate2021feasibility}
G.~Khandate, M.~Haas-Heger, and M.~Ciocarlie, ``On the feasibility of learning
  finger-gaiting in-hand manipulation with intrinsic sensing,'' \emph{arXiv
  preprint arXiv:2109.12720}, 2021.

\bibitem{sundaralingam2021hand}
B.~Sundaralingam and T.~Hermans, ``In-hand object-dynamics inference using
  tactile fingertips,'' \emph{IEEE Transactions on Robotics}, vol.~37, no.~4,
  pp. 1115--1126, 2021.

\bibitem{charlesworth2021solving}
H.~J. Charlesworth and G.~Montana, ``Solving challenging dexterous manipulation
  tasks with trajectory optimisation and reinforcement learning,'' in
  \emph{International Conference on Machine Learning}.\hskip 1em plus 0.5em
  minus 0.4em\relax PMLR, 2021, pp. 1496--1506.

\bibitem{ibarz2021train}
J.~Ibarz, J.~Tan, C.~Finn, M.~Kalakrishnan, P.~Pastor, and S.~Levine, ``How to
  train your robot with deep reinforcement learning: lessons we have learned,''
  \emph{The International Journal of Robotics Research}, vol.~40, no. 4-5, pp.
  698--721, 2021.

\bibitem{gupta2021reset}
A.~Gupta, J.~Yu, T.~Z. Zhao, V.~Kumar, A.~Rovinsky, K.~Xu, T.~Devlin, and
  S.~Levine, ``Reset-free reinforcement learning via multi-task learning:
  Learning dexterous manipulation behaviors without human intervention,'' in
  \emph{2021 IEEE International Conference on Robotics and Automation
  (ICRA)}.\hskip 1em plus 0.5em minus 0.4em\relax IEEE, 2021, pp. 6664--6671.

\bibitem{chen2021simple}
T.~Chen, J.~Xu, and P.~Agrawal, ``A simple method for complex in-hand
  manipulation,'' in \emph{5th Annual Conference on Robot Learning}, vol.~3,
  2021.

\bibitem{huang2021generalization}
W.~Huang, I.~Mordatch, P.~Abbeel, and D.~Pathak, ``Generalization in dexterous
  manipulation via geometry-aware multi-task learning,'' \emph{arXiv preprint
  arXiv:2111.03062}, 2021.

\bibitem{chen2022system}
T.~Chen, J.~Xu, and P.~Agrawal, ``A system for general in-hand object
  re-orientation,'' in \emph{Conference on Robot Learning}.\hskip 1em plus
  0.5em minus 0.4em\relax PMLR, 2022, pp. 297--307.

\bibitem{taylor2009transfer}
M.~E. Taylor and P.~Stone, ``Transfer learning for reinforcement learning
  domains: A survey,'' \emph{Journal of Machine Learning Research}, vol.~10,
  no.~7, 2009.

\bibitem{yang2020transfer}
Q.~Yang, Y.~Zhang, W.~Dai, and S.~J. Pan, \emph{Transfer learning}.\hskip 1em
  plus 0.5em minus 0.4em\relax Cambridge University Press, 2020.

\bibitem{han2022transfer}
H.~Han, H.~Liu, C.~Yang, and J.~Qiao, ``Transfer learning algorithm with
  knowledge division level,'' \emph{IEEE Transactions on Neural Networks and
  Learning Systems}, pp. 1--15, 2022, doi:{\color{blue}
  \href{https://doi.org/10.1109/TNNLS.2022.3151646}{10.1109/TNNLS.2022.3151646}}.

\bibitem{Haarnoja2017ReinforcementLW}
T.~Haarnoja, H.~Tang, P.~Abbeel, and S.~Levine, ``Reinforcement learning with
  deep energy-based policies,'' in \emph{ICML}, 2017.

\bibitem{Florensa2017StochasticNN}
C.~Florensa, Y.~Duan, and P.~Abbeel, ``Stochastic neural networks for
  hierarchical reinforcement learning,'' \emph{ArXiv}, vol. abs/1704.03012,
  2017.

\bibitem{Kumar2020OneSI}
S.~Kumar, A.~Kumar, S.~Levine, and C.~Finn, ``One solution is not all you need:
  Few-shot extrapolation via structured maxent rl,'' \emph{ArXiv}, vol.
  abs/2010.14484, 2020.

\bibitem{Tzeng2016AdaptingDV}
E.~Tzeng, C.~Devin, J.~Hoffman, C.~Finn, P.~Abbeel, S.~Levine, K.~Saenko, and
  T.~Darrell, ``Adapting deep visuomotor representations with weak pairwise
  constraints,'' in \emph{WAFR}, 2016.

\bibitem{Eysenbach2021OffDynamicsRL}
B.~Eysenbach, S.~Asawa, S.~Chaudhari, R.~Salakhutdinov, and S.~Levine,
  ``Off-dynamics reinforcement learning: Training for transfer with domain
  classifiers,'' \emph{ArXiv}, vol. abs/2006.13916, 2021.

\bibitem{chen2022domain}
J.~Chen, X.~Wu, L.~Duan, and S.~Gao, ``Domain adversarial reinforcement
  learning for partial domain adaptation,'' \emph{IEEE Transactions on Neural
  Networks and Learning Systems}, vol.~33, no.~2, pp. 539--553, 2022.

\bibitem{tobin2017domain}
J.~Tobin, R.~Fong, A.~Ray, J.~Schneider, W.~Zaremba, and P.~Abbeel, ``Domain
  randomization for transferring deep neural networks from simulation to the
  real world,'' in \emph{2017 IEEE/RSJ international conference on intelligent
  robots and systems (IROS)}.\hskip 1em plus 0.5em minus 0.4em\relax IEEE,
  2017, pp. 23--30.

\bibitem{peng2018sim}
X.~B. Peng, M.~Andrychowicz, W.~Zaremba, and P.~Abbeel, ``Sim-to-real transfer
  of robotic control with dynamics randomization,'' in \emph{2018 IEEE
  international conference on robotics and automation (ICRA)}.\hskip 1em plus
  0.5em minus 0.4em\relax IEEE, 2018, pp. 3803--3810.

\bibitem{Rusu2016PolicyD}
A.~A. Rusu, S.~G. Colmenarejo, Çaglar G{\"u}lçehre, G.~Desjardins,
  J.~Kirkpatrick, R.~Pascanu, V.~Mnih, K.~Kavukcuoglu, and R.~Hadsell, ``Policy
  distillation,'' \emph{CoRR}, vol. abs/1511.06295, 2016.

\bibitem{Berseth2018ProgressiveRL}
G.~Berseth, K.~Xie, P.~Cernek, and M.~van~de Panne, ``Progressive reinforcement
  learning with distillation for multi-skilled motion control,'' \emph{ArXiv},
  vol. abs/1802.04765, 2018.

\bibitem{Gimelfarb2021ContextualPT}
M.~Gimelfarb, S.~Sanner, and C.-G. Lee, ``Contextual policy transfer in
  reinforcement learning domains via deep mixtures-of-experts,'' in \emph{UAI},
  2021.

\bibitem{sun2021model}
Y.~Sun, K.~Zhang, and C.~Sun, ``Model-based transfer reinforcement learning
  based on graphical model representations,'' \emph{IEEE Transactions on Neural
  Networks and Learning Systems}, pp. 1--14, 2021, doi:{\color{blue}
  \href{https://doi.org/10.1109/TNNLS.2021.3107375}{10.1109/TNNLS.2021.3107375}}.

\bibitem{Sasso2022MultiSourceTL}
R.~Sasso, M.~Sabatelli, and M.~A. Wiering, ``Multi-source transfer learning for
  deep model-based reinforcement learning,'' \emph{ArXiv}, vol. abs/2205.14410,
  2022.

\bibitem{rusu2016progressive}
A.~A. Rusu, N.~C. Rabinowitz, G.~Desjardins, H.~Soyer, J.~Kirkpatrick,
  K.~Kavukcuoglu, R.~Pascanu, and R.~Hadsell, ``Progressive neural networks,''
  \emph{arXiv preprint arXiv:1606.04671}, 2016.

\bibitem{rusu2017sim}
A.~A. Rusu, M.~Ve{\v{c}}er{\'\i}k, T.~Roth{\"o}rl, N.~Heess, R.~Pascanu, and
  R.~Hadsell, ``Sim-to-real robot learning from pixels with progressive nets,''
  in \emph{Conference on Robot Learning}.\hskip 1em plus 0.5em minus
  0.4em\relax PMLR, 2017, pp. 262--270.

\bibitem{Barreto2018TransferID}
A.~Barreto, D.~Borsa, J.~Quan, T.~Schaul, D.~Silver, M.~Hessel, D.~J.
  Mankowitz, A.~Z{\'i}dek, and R.~Munos, ``Transfer in deep reinforcement
  learning using successor features and generalised policy improvement,'' in
  \emph{ICML}, 2018.

\bibitem{Abdolshah2021ANR}
M.~Abdolshah, H.~Le, T.~G. Karimpanal, S.~Gupta, S.~Rana, and S.~Venkatesh, ``A
  new representation of successor features for transfer across dissimilar
  environments,'' \emph{ArXiv}, vol. abs/2107.08426, 2021.

\bibitem{allshire2021transferring}
A.~Allshire, M.~Mittal, V.~Lodaya, V.~Makoviychuk, D.~Makoviichuk, F.~Widmaier,
  M.~W{\"u}thrich, S.~Bauer, A.~Handa, and A.~Garg, ``Transferring dexterous
  manipulation from gpu simulation to a remote real-world trifinger,''
  \emph{arXiv preprint arXiv:2108.09779}, 2021.

\bibitem{lin1992self}
L.-J. Lin, ``Self-improving reactive agents based on reinforcement learning,
  planning and teaching,'' \emph{Machine learning}, vol.~8, no. 3-4, pp.
  293--321, 1992.

\bibitem{schaul2015prioritized}
T.~Schaul, J.~Quan, I.~Antonoglou, and D.~Silver, ``Prioritized experience
  replay,'' \emph{arXiv preprint arXiv:1511.05952}, 2015.

\bibitem{de2018experience}
T.~De~Bruin, J.~Kober, K.~Tuyls, and R.~Babuska, ``Experience selection in deep
  reinforcement learning for control,'' \emph{Journal of Machine Learning
  Research}, vol.~19, 2018.

\bibitem{fedus2020revisiting}
W.~Fedus, P.~Ramachandran, R.~Agarwal, Y.~Bengio, H.~Larochelle, M.~Rowland,
  and W.~Dabney, ``Revisiting fundamentals of experience replay,'' in
  \emph{International Conference on Machine Learning}.\hskip 1em plus 0.5em
  minus 0.4em\relax PMLR, 2020, pp. 3061--3071.

\bibitem{yarats2022don}
D.~Yarats, D.~Brandfonbrener, H.~Liu, M.~Laskin, P.~Abbeel, A.~Lazaric, and
  L.~Pinto, ``Don't change the algorithm, change the data: Exploratory data for
  offline reinforcement learning,'' \emph{arXiv preprint arXiv:2201.13425},
  2022.

\bibitem{lazaric2008transfer}
A.~Lazaric, M.~Restelli, and A.~Bonarini, ``Transfer of samples in batch
  reinforcement learning,'' in \emph{Proceedings of the 25th international
  conference on Machine learning}, 2008, pp. 544--551.

\bibitem{tirinzoni2018importance}
A.~Tirinzoni, A.~Sessa, M.~Pirotta, and M.~Restelli, ``Importance weighted
  transfer of samples in reinforcement learning,'' in \emph{International
  Conference on Machine Learning}.\hskip 1em plus 0.5em minus 0.4em\relax PMLR,
  2018, pp. 4936--4945.

\bibitem{tirinzoni2019transfer}
A.~Tirinzoni, M.~Salvini, and M.~Restelli, ``Transfer of samples in policy
  search via multiple importance sampling,'' in \emph{International Conference
  on Machine Learning}.\hskip 1em plus 0.5em minus 0.4em\relax PMLR, 2019, pp.
  6264--6274.

\bibitem{yin2017knowledge}
H.~Yin and S.~J. Pan, ``Knowledge transfer for deep reinforcement learning with
  hierarchical experience replay,'' in \emph{Thirty-First AAAI conference on
  artificial intelligence}, 2017.

\bibitem{Kalashnikov2021MTOptCM}
D.~Kalashnikov, J.~Varley, Y.~Chebotar, B.~Swanson, R.~Jonschkowski, C.~Finn,
  S.~Levine, and K.~Hausman, ``Mt-opt: Continuous multi-task robotic
  reinforcement learning at scale,'' \emph{ArXiv}, vol. abs/2104.08212, 2021.

\bibitem{wiering2012reinforcement}
M.~A. Wiering and M.~Van~Otterlo, ``Reinforcement learning,'' \emph{Adaptation,
  learning, and optimization}, vol.~12, no.~3, p. 729, 2012.

\bibitem{pan2018multisource}
J.~Pan, X.~Wang, Y.~Cheng, and Q.~Yu, ``Multisource transfer double dqn based
  on actor learning,'' \emph{IEEE transactions on neural networks and learning
  systems}, vol.~29, no.~6, pp. 2227--2238, 2018.

\bibitem{wolpert2000computational}
D.~M. Wolpert and Z.~Ghahramani, ``Computational principles of movement
  neuroscience,'' \emph{Nature neuroscience}, vol.~3, no.~11, pp. 1212--1217,
  2000.

\bibitem{flash1985coordination}
T.~Flash and N.~Hogan, ``The coordination of arm movements: an experimentally
  confirmed mathematical model,'' \emph{Journal of neuroscience}, vol.~5,
  no.~7, pp. 1688--1703, 1985.

\bibitem{daley2019reconciling}
B.~Daley and C.~Amato, ``Reconciling $\lambda$-returns with experience
  replay,'' \emph{Advances in Neural Information Processing Systems}, vol.~32,
  2019.

\end{thebibliography}

\vfill

\end{document}